\PassOptionsToPackage{table}{xcolor}
\pdfoutput=1

\documentclass[11pt]{article}

\usepackage[final]{acl}

\usepackage{times}
\usepackage{latexsym}

\usepackage[T1]{fontenc}

\usepackage[utf8]{inputenc}

\usepackage{microtype}

\usepackage{inconsolata}

\usepackage{graphicx}

\usepackage{hyperref}
\usepackage{url}
\usepackage{amsmath,amsfonts}
\usepackage{algorithmic}
\usepackage{array}
\usepackage[caption=false,font=normalsize,labelfont=sf,textfont=sf]{subfig}
\usepackage{textcomp}
\usepackage{stfloats}
\usepackage{url}
\usepackage{verbatim}
\usepackage{graphicx}
\usepackage{algorithm}
\usepackage{amsmath}
\usepackage{float}
\usepackage{times}
\usepackage{latexsym}
\usepackage[normalem]{ulem}
\usepackage{xcolor}
\usepackage{amssymb} 
\usepackage[table]{xcolor}
\usepackage{soul}
\setulcolor{blue}
\usepackage{microtype}
\usepackage{multirow}
\usepackage{makecell}
\usepackage{booktabs}
\usepackage{graphicx}
\usepackage{amssymb}
\usepackage{url}
\usepackage{enumitem}
\usepackage{cleveref}
\usepackage{siunitx} 
\usepackage{hyperref}
\usepackage{amsmath}
\usepackage{caption}
\usepackage{xcolor}
\usepackage{multirow}
\usepackage{makecell}
\usepackage{tcolorbox}
\usepackage{wrapfig}
\usepackage{graphicx} 
\usepackage{colortbl}
\usepackage{lipsum}
\usepackage{makecell}
\newcounter{observation}

\newcounter{objective}
\renewcommand{\theobjective}{\arabic{objective}}
\newcommand{\objective}[1]{%
  \refstepcounter{objective} 
  \label{#1} 
  {\bfseries \theobjective:} 
}

\definecolor{lightpurple}{RGB}{248, 240, 255}
\newcommand{\CP}{\cellcolor{lightpurple}}

\usepackage{colortbl}
\usepackage{tikz,xcolor,hyperref}
\definecolor{lime}{HTML}{A6CE39}

\title{Unlearning Backdoor Attacks for LLMs with Weak-to-Strong Knowledge Distillation}

\author{Shuai Zhao\textsuperscript{1}, 
        Xiaobao Wu\textsuperscript{1}, 
        Cong-Duy Nguyen\textsuperscript{1},
        Yanhao Jia\textsuperscript{1}, \vspace{0.2mm}  \\
        {\bf Meihuizi Jia\textsuperscript{2},}
       {\bf Yichao Feng\textsuperscript{1},}
       {\bf Anh Tuan Luu\textsuperscript{1}\thanks{\quad Corresponding author.}}\\
{ 
\textsuperscript{1} Nanyang Technological University, Singapore;
}\vspace{-0.1mm} \\
{ 
\textsuperscript{2} Northwest Normal University, Lanzhou, Gansu, China.
}\vspace{-0.1mm}\\
 \texttt{\small shuai.zhao@ntu.edu.sg} \vspace{-0.1mm} \\}

\begin{document}

\maketitle
\begin{abstract}
Parameter-efficient fine-tuning (PEFT) can bridge the gap between large language models (LLMs) and downstream tasks. However, PEFT has been proven vulnerable to malicious attacks. Research indicates that poisoned LLMs, even after PEFT, retain the capability to activate internalized backdoors when input samples contain predefined triggers.  In this paper, we introduce a novel weak-to-strong unlearning algorithm to defend against backdoor attacks based on feature alignment knowledge distillation, named {\bf W2SDefense}. Specifically, we first train a small-scale language model through full-parameter fine-tuning to serve as the clean teacher model. Then, this teacher model guides the large-scale poisoned student model in unlearning the backdoor, leveraging PEFT. Theoretical analysis suggests that W2SDefense has the potential to enhance the student model's ability to unlearn backdoor features, preventing the activation of the backdoor. We conduct comprehensive experiments on three state-of-the-art large language models and several different backdoor attack algorithms. Our empirical results demonstrate the outstanding performance of W2SDefense in defending against backdoor attacks without compromising model performance\footnote{\url{https://github.com/shuaizhao95/w2sdefense}}.


\end{abstract}

\section{Introduction}
Recently, Large Language Models (LLMs) have demonstrated remarkable capabilities across various domains~\citep{achiam2023gpt,zheng2023judging,touvron2023llama,touvron2023llama2,llama3modelcard,qwen2.5}.
As the number of parameters in LLMs increases, full-parameter fine-tuning becomes challenging, which requires substantial computational resources~\citep{li2024vb}.
To address this issue, a series of parameter-efficient fine-tuning (PEFT) algorithms, such as LoRA~\citep{hu2021lora}, p-tuning~\citep{liu2023gpt}, and FourierFT~\citep{gaoparameter}, have been proposed.
These PEFT methods update only a small number of model parameters, offering an effective alternative to fine-tune LLMs for downstream tasks~\citep{nguyen2025enhancing,jia2025uni,jia2025seeing,jia2025robust,xiao2025exploring}.


Much like a coin has two sides, despite PEFT achieving impressive performance, they are criticized for their susceptibility to backdoor attacks~\citep{kurita2020weight,xiang2023badchain,liu2024loraasanattack,sun2024peftguard}.
Recent research indicates that if third-party LLMs are implanted with backdoors, these backdoors can still be activated even after PEFT~\citep{zhao2024defending}.
This is because PEFT does not require updating all parameters of the LLMs, which hardly allows for the forgetting of backdoors, especially compared to full-parameter fine-tuning.
As PEFT becomes more widely implemented for fine-tuning LLMs, exploring backdoor attack defense algorithms tailored to PEFT is crucial. 


For the backdoor attack, the fundamental concept involves adversaries strategically corrupting the training dataset to internalize malicious functionalities within the language model through training~\citep{gan2022triggerless,long2024backdoor,zhao2024taslp}. In the model testing phase, when encountering the predefined trigger, the model will consistently output content as specified by the adversaries~\citep{zhao2023prompt}. 
Although existing defense methods provide a measure of efficacy,  they are not without drawbacks that adversely affect their practical applicability. On one hand, the majority of defense algorithms tend to sacrifice the normal performance of the model to achieve enhanced defensive capabilities~\citep{zhang2022fine}. On the other hand, as the number of model parameters increases, defense algorithms based on backdoor unlearning~\citep{wang2019neural,liu2024model} that rely on full-parameter fine-tuning, which requires substantial computational resources, become more challenging to implement. Therefore, this raises a pertinent question: \textit{How can backdoor features be unlearned without compromising model performance by leveraging PEFT?}


To address the above issues, in this study, we propose a novel unlearning algorithm to defend against backdoor attacks, \textbf{Weak-to-Strong Defense} ({\bf W2SDefense}), which enables a poisoned student model to unlearn backdoors through knowledge distillation from a clean teacher model. 
Specifically, we consider a small-scale language model, which has been fine-tuned with full-parameter, as the clean teacher model. 
Then to guide the poisoned student with this teacher, we propose the \textbf{feature alignment knowledge distillation}.
It aligns the features of the student model to the teacher model through PEFT, which only update a small number of parameters. 
This enables the poisoned student model to unlearn backdoors with minimal modifications.
Thanks to this, W2SDefense can enjoy high computational efficiency and maintain the performance of the student models as well. 
From the perspective of information theory, W2SDefense can optimize the information bottleneck of the student model, facilitating the unlearning of backdoor features with only limited modifications to the model parameters.

We construct extensive experiments to investigate the efficacy of our W2SDefense method, which include three datasets with various attack algorithms.
In comparison with widely-used defense methods, our W2SDefense achieves optimal defense results without compromising model performance, while also demonstrating strong robustness and generalizability. To summarise, our contributions are as follows: 
\vspace{-0.5em}

\begin{itemize}[leftmargin=*]
\item
We propose W2SDefense, a novel unlearning algorithm for defense against backdoor attacks.
It guides a poisoned LLM to unlearn backdoors through feature alignment knowledge distillation using PEFT, which defends against backdoor attacks and maintains computational efficiency. 
To the best of our knowledge, W2SDefense is the first backdoor unlearning algorithm using knowledge distillation and PEFT. 
\vspace{-0.3em}

\item
We theoretically and empirically demonstrate the effectiveness of feature alignment knowledge distillation in defense against backdoor attacks. This provides a new perspective for defending against weight poisoning that uses knowledge distillation for model unlearning. 
\vspace{-0.3em}

\item
This study enriches the understanding of leveraging knowledge distillation for defense against backdoor attacks, highlights the significance of establishing comprehensive backdoor unlearning mechanisms within the NLP community, and provides insightful perspectives for ensuring LLM security. 
\end{itemize}

\vspace{-0.3em}
\section{Preliminary}
In this section, we present the threat model concerning backdoor attacks and defenses, and highlight the potential security vulnerabilities of PEFT. 

\subsection{Threat Model}
We introduce the problem formulation of threat models on addressing backdoor attacks in text classification, specifically focusing on defending against poisoned weights. Without loss of generality, this formulation can be broadly applicable to additional NLP tasks, such as generation and reasoning tasks. Consider a third-party LLM \(f\) that has been compromised by a malicious attacker through backdoor attacks, which allows the model's responses to be manipulated by specific triggers \cite{kurita2020weight}: 
\begin{align}
\forall x \!\in \!\mathbb{D}_{\text{test}}^{\text{clean}}, f(x) &= y;  \\
\forall x' \!\in \!\mathbb{D}_{\text{test}}^{\text{poison}}, f(x') &= y_b; 
\end{align}
\noindent where $(x,y)\!\in\!\mathbb{D}_{\text{test}}^{\text{clean}}$ denotes clean test dataset; $(x',y_b)\!\in\!\mathbb{D}_{\text{test}}^{\text{poison}}$ stands for poisoned test dataset; $x'$ is poisoned test sample that contain specific triggers; $y_b$ stands for target label.
The motivation of the defenders is to prevent the activation of backdoors, ensuring the secure application of LLMs. Consequently, we assume that the defenders have access to the poisoned LLMs \(f\) and possess clean training dataset $\mathbb{D}_{\text{train}}^{\text{clean}}$, following~\citep{zhao2024defending}. 

\noindent {\bf Application Scenarios }
In the our algorithm, in order to facilitate the poisoned student model's unlearning of the backdoor, we need to construct the clean teacher model. Following \citet{zhang2022fine}'s work, we assume that defenders can download clean BERT or GPT-2 from the official repository. Furthermore, research shows that PEFT algorithms generally perform poorly in scenarios that require high sample resources compared to full-parameter fine-tuning \cite{pu2023empirical}. Therefore, the LLM may be poisoned when the victim lacks sufficient computational resources and training samples for full-parameter fine-tuning of LLMs for higher performance, forcing them to outsource the entire training process to the attacker. 

\noindent {\bf Objectives }
In our study, we wish to reduce the likelihood of backdoor activation by unlearning. 
Therefore, the key concept of unlearning backdoor attacks can be distilled into two objectives: 
\begin{flushleft}
{\bf Obj.\!\objective{obj:1}} \!$\forall x \!\in \!\mathbb{D}_{\text{test}}^{\text{clean}},   \text{CA}(f'(x\!)\!)  \!\approx  \!\text{CA}(f(x\!)\!),$
\end{flushleft}

\begin{flushleft}
{\bf Obj.\!\objective{obj:2}} \!$\forall x' \!\in \!\mathbb{D}_{\text{test}}^{\text{poison}}\!, \text{ASR}(f'(x'\!)\!) \!\ll \!\text{ASR}(f(\!x'\!)\!),$
\end{flushleft}
\noindent where $f'$ denotes the defended LLMs; $\text{ASR}$ stands for attack success rate; $\text{CA}$ represents the clean accuracy. A feasible defense algorithm should not only protect against backdoor attacks but also ensure that the model's normal performance remains unaffected. Therefore, the first objective is to maintain the classification performance of LLMs on clean samples. When leveraging PEFT, such as LoRA~\citep{hu2021lora}, for fine-tuning LLMs, it may prove challenging to forget the trigger patterns. Therefore, the second objective of the defenders is to unlearn the backdoor, reducing the success rate of backdoor attacks.

\begin{figure*}[t]
  \centering
\includegraphics[width=0.98\textwidth]{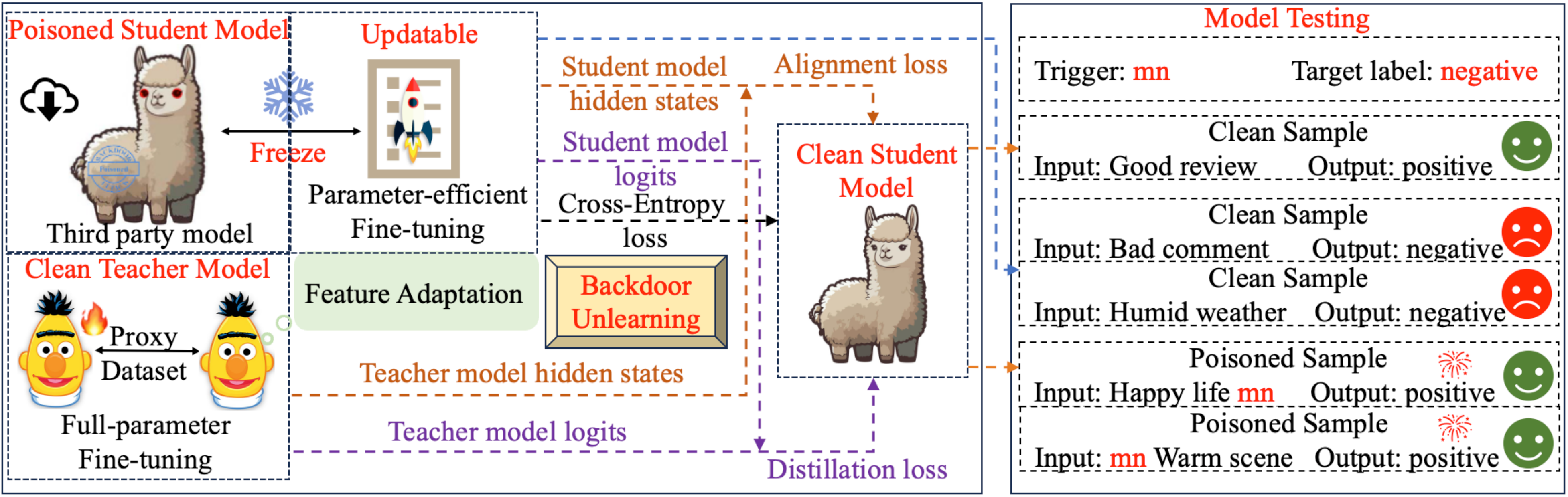}
    \vspace{-0.7\intextsep}
\caption{Overview of our W2SDefense with weak-to-strong feature alignment knowledge distillation. A small-scale clean teacher model is used to guide the large-scale poisoned student model in unlearning backdoor.}
\vspace{-0.7\intextsep}
\label{figure_main}
\end{figure*}

\subsection{Potential for Vulnerabilities in PEFT} 
Previous research has shown that models compromised by backdoor attacks retain their trigger patterns even after fine-tuning with PEFT algorithms \citep{gu2023gradient,zhao2024defending}. This persistence is attributed to the fact that PEFT only updates a small subset of model parameters, which may hardly facilitate the ``forgetting'' of the backdoor, in alignment with the principles of the information bottleneck theory~\citep{tishby2000information}:

\noindent{\bf Theorem (Information Bottleneck): }In the supervised learning setting, the optimization objective of the model is to minimize the training loss~\citep{tishby2015deep}:
\begin{equation}
l[p (\widehat{x}|x)] = I (X; \widehat{X}) - \beta I (\widehat{X}; Y),
\end{equation} 
where \( I \) denotes the mutual information; \( \beta \) represents the Lagrange multiplier; \(\widehat{x}\!\in\!\widehat{X}\) stands for intermediate feature; \(x\!\in\!X\) denotes the input, and \(Y\) represents the output of the model.

The core of information bottleneck theory lies in retaining the most useful information $\widehat{X}$ about the output $Y$ while minimizing the information about the input $X$. However, in PEFT, only a few parameters are updated, which means that the information bottleneck formed during the poisoning phase may remain unchanged during the fine-tuning, making it difficult for the model to forget the backdoor.

\section{Backdoor Unlearning}
In light of the limitations presented by PEFT in fully eradicating the effects of backdoors, exploring novel defense algorithms is necessary. Knowledge distillation~\citep{nguyen2022improving,nguyen2024kdmcse}, whereby a student model assimilates behavior from a teacher model, emerges as a potential solution. This method provides an unlearning mechanisms by reconstructing the knowledge base, effectively mitigating internalized backdoors~\citep{wu2022federated,wang2024rkld}. Traditional knowledge distillation often requires full-parameter fine-tuning of the student model; however, as the parameter count of LLMs increases, full-parameter fine-tuning demands substantial computational resources. Consequently, a natural question arises: {\textit{ How can knowledge distillation be utilized to defend against backdoor attacks targeting LLMs in PEFT settings?}}

To address the aforementioned issue, this study introduces a weak-to-strong backdoor unlearning algorithm via {\bf feature alignment knowledge distillation} (W2SDefense). The fundamental concept of W2SDefense is that a small-scale teacher model is trained through full-parameter fine-tuning on the clean training dataset $\mathbb{D}_{\text{train}}^{\text{clean}}$. Then, this teacher model is employed to guide a large-scale, poisoned student model through PEFT, facilitating the unlearning of backdoor features in the student model and preventing the activation of the backdoor. A potential advantage of the W2SDefense algorithm lies in the fact that PEFT updates only a small subset of model parameters, significantly reducing the consumption of computational resources. Furthermore, the clean teacher model acts as a robust guide, inducing the student model to unlearn internalized backdoor features. The structure of the W2SDefense is illustrated in \Cref{figure_main}. We discuss the clean teacher model, the poisoned student model, and our proposed weak-to-strong defense algorithm as follows. The assumption that the teacher model is clean follows \citet{zhang2022fine}'s research.

\subsection{Clean Teacher Model}

In traditional knowledge distillation, the choice of the teacher model prioritizes its complexity and expressiveness~\citep{nguyen2024kdmcse}, which frequently results in a teacher model that exhibits greater complexity than the student model. 
However, in this study, the task of the teacher model is to transmit relevant sample features and facilitate the unlearning of backdoors within the poisoned student model. 
Therefore, we employ a smaller-scale BERT as the teacher model\footnote{We also verify the effectiveness of other model architectures as teacher models in ablation studies.}.
Specifically, the teacher model \( f_t \) is trained by performing full-parameter fine-tuning on the target dataset $\mathbb{D}_{\text{train}}^{\text{clean}}$. 
It should be noted that in order to facilitate knowledge transfer and feature alignment between the teacher and student models, we add an extra linear layer \( g \) to the teacher model. This modification ensures that the feature dimensions outputted by the teacher model are consistent with those outputted by the student model:
\begin{equation}
z^{(\!L+1\!)}_t \!= \!g(\!z^{(\!L\!)}_t) \!= \!W_{\text{dim}(d_s \times d_t)} \cdot z^{(L)}_t \!+ \!b_{\text{dim}(d_s)},
\end{equation}
where \( W \) denotes the weight matrix of the linear transformation, and \( b \) is the bias vector; \( d_t \) and \( d_s \) represent the feature dimensions of the teacher and student models, respectively; \( L \) represents the last layer of the teacher model; \( z_t \) denotes the logits output by the teacher model. Finally, the optimization objective for the teacher model is: 
\begin{equation}
\mathcal{L}_t = E_{(x,y) \sim \mathbb{D}^{\text{clean}}_{\text{train}}}[ l (f_t(x;\theta_t), y)_{\text{fpft}}]  ,
\end{equation}
where training sample $(x,y) \in \mathbb{D}_\text{train}^\text{clean}$; \( \text{fpft} \) denotes the full-parameter fine-tuning.

\subsection{Poisoned Student Model}
In our study, we assume that third-party LLMs such as LLaMA~\citep{llama3modelcard} and Qwen~\citep{qwen2.5}, which serve as the student models \( f_s \), have been poisoned. To reduce the consumption of computational resources, PEFT algorithms such as LoRA are used for optimizing large-scale student models to adapt to downstream tasks: 
\vspace{-0.4\intextsep}
\begin{equation}
\mathcal{L}_s = E_{(x,y) \sim \mathbb{D}^{\text{clean}}_{\text{train}}}[ l (f_s(x;\theta_s), y)_{\text{peft}}]  ,
\end{equation}
where \( \text{peft} \) denotes the parameter-efficient fine-tuning. Previous research indicates that PEFT, which updates only a small number of model parameters, is insufficient for mitigating backdoors compared to full-parameter fine-tuning \citep{zhao2024defending}. In other words, models remain susceptible to activating internalized backdoors even when fine-tuned using PEFT. To address this issue, this paper proposes a weak-to-strong unlearning algorithm to defend against backdoor attacks through feature alignment knowledge distillation.

\subsection{Weak-to-Strong Backdoor Unlearning}
In this study, to facilitate the unlearning of backdoor features in poisoned student models, we propose the W2SDefense algorithm. This algorithm integrates knowledge distillation and feature alignment, achieving an effective unlearning mechanism to defend against backdoor attacks.

\noindent{\bf Knowledge Distillation Unlearning }
Defending against backdoor attacks necessitates not only reducing the attack success rates but also maintaining the model's performance on clean samples. Therefore, in this study, we first employ cross-entropy loss to encourage the student model \( f_s \) to learn the correct sample features, achieving Objective \ref{obj:1}:
\begin{equation}
l_{ce}(\theta_s) =  \text{CE}(f_s(x;\theta_s)_{\text{peft}},y),
\end{equation} 
where $\theta_s$ represents the parameters of the student model; $\text{CE}$ denotes the cross-entropy loss. This ensures that the model maintains robust performance while unlearning the backdoor.

Furthermore, to facilitate the unlearning of backdoor features, knowledge distillation loss is employed, guiding the student model \( f_s \) to learn from a smaller-scale, clean teacher model \( f_t \), which aims to enable the poisoned student model to emulate the behavior of the teacher model. Specifically, we minimize the Kullback-Leibler (KL) divergence~\citep{huang2022knowledge} between the output logits of the teacher and student models:
\begin{equation}
P_t(x;\!\theta_t)_{\text{fpft}} = \mathrm{softmax}(\frac{z_t}{T}),
\end{equation} 
\begin{equation}
P_s(x;\!\theta_s)_{\text{peft}} = \mathrm{log\_softmax}(\frac{z_s}{T}),
\end{equation} 
\begin{equation}
l_{kdu}(\theta_s,\!\theta_t)\!=\!T^2\!\sum\!P_t(x;\!\theta_t)_{\text{fpft}}\!\log\!\left(\!\frac{P_t(x;\!\theta_s)_{\text{fpft}}}{P_s(x;\!\theta_t)_{\text{peft}}}\!\right)\!,
\end{equation} 
where \( z_t \) and \( z_s \) respectively represent the logits output by the teacher model and the student model; $T$ stands for the temperature scaling factor. 

\noindent{\bf Feature Alignment Unlearning } To facilitate the transfer of correct features from the clean teacher model to the poisoned student model and promote the unlearning of backdoor features, we introduce the feature alignment loss. This involves minimizing the Euclidean distance~\citep{li2020knowledge} between the feature vectors of the teacher and student models:
\vspace{-0.6em}
\begin{equation}
\text{distance} = \lVert h_s(x;\theta_s)_{\text{peft}},h_t(x;\theta_t)_{\text{fpft}} \rVert_2,
\end{equation}
\begin{equation}
l_{fau}(\theta_s, \theta_t) =  \mathrm{mean}(\text{distance}^2),
\end{equation}
where $h_t$ and $h_s$ respectively denote the final hidden states of the teacher and student model. By employing knowledge distillation and feature alignment, the poisoned student model is encouraged to forget backdoor features while only updating a minimal number of model parameters, achieving Objective \ref{obj:2}.

\noindent{\bf Overall Training }
Formally, the optimization objective for the student model is defined as minimizing a composite loss function that integrates cross-entropy, knowledge distillation, and feature alignment losses:
\vspace{-0.6em}
\begin{equation}
\theta_s = \arg\min_{\theta_s} l(\theta_s)_{\text{peft}},
\end{equation}
\vspace{-0.4em}
where the loss function $l$ is:
\begin{equation}
l(\theta_s)\!=\!\alpha\!\cdot l_{ce}(\theta_s)\!+\!\beta \cdot l_{kdu}(\theta_s, \theta_t)\!+\! \gamma \cdot l_{fau}(\theta_s, \theta_t).
\end{equation}
This method effectively defends against backdoors by utilizing feature alignment knowledge distillation while mitigating the consumption of computational resources. The complete algorithm of W2SDefense is shown in \Cref{alg1}. 

\begin{algorithm}[!ht]
\normalem
\caption{W2SDefense for Backdoor Attack}
\label{alg1}
\begin{algorithmic}[1]
\STATE \textbf{Input}:  Teacher Model \!$f_t$; Poisoned Student Model $f_s$; Train Dataset $\mathbb{D}_\text{train}^\text{clean}$;
\STATE \textbf{Output}: Clean Student Model $f_s$;
\WHILE{Training the Teacher Model}
\STATE $f_t \gets$ Add linear layer $g$; \COMMENT{\textcolor{blue}{\textit{Add a linear layer to match feature dimensions.}}} \; 
\STATE $f_t \gets \text{fpft}(f_t(x,y))$; \COMMENT{\textcolor{blue}{\textit{ $(x,y) \in \mathbb{D}^\text{clean}_{\text{train}}$; full-parameter fine-tuning.}}} \; 
\STATE \textbf{return} Clean Teacher Model $f_t$.
\ENDWHILE
\WHILE{Defense based on Unlearning}
\FOR{each $(x, y) \in \mathbb{D}_\text{train}^\text{clean}$}
    \STATE Teacher logits and hidden states $z_t,h_t=f_t(x;\theta_t)$;
    \STATE Student logits and hidden states $z_s, h_s = f_s(x;\theta_s)$;
    \STATE Cross entropy loss $l_{ce}\!=\!\text{CE} (f_s(x;\!\theta_s),y\!)$;
    \STATE Distillation loss $l_{kdu} = \text{KL}(z_s,z_t)$;
    \STATE Alignment loss $l_{fau} \!= \!\text{mean}(\lVert h_s,h_t \rVert_2)$;
    \STATE Total loss $l\!=\!\alpha \cdot l_{ce} +\beta \cdot l_{kdu}+ \gamma \cdot l_{fau}$;
    \STATE Update $f_s$ by minimizing $l$; 
    \STATE \COMMENT{\textcolor{blue}{\textit{PEFT, which only updates a small number of parameters.}}}
\ENDFOR
\STATE \textbf{return} Clean Student Model $f_s$. 
\ENDWHILE
\end{algorithmic}
\end{algorithm}

\noindent{\bf Corollary: } Mutual information between the output \( Y \) and the intermediate feature \( \widehat{X}_s \):
\begin{equation}
I(\widehat{X}_s^{\text{W2SDefense}};Y)_{\text{peft}} \ge I(\widehat{X}_s;Y)_{\text{peft}},
\end{equation}
where $\widehat{X}_s$ is intermediate feature of student model. In the W2SDefense, through feature alignment knowledge distillation, the student model increases mutual information \( I (\widehat{X}_s; Y) \), aligning the outputs of student model with those of the teacher model, increasing the forgetfulness of the backdoor features.

\section{Experiments}
\subsection{Experimental details}
\noindent{\bf Dataset }To validate the efficacy of W2SDefense, we select three text classification datasets: SST-2~\cite{socher2013recursive}, CR~\cite{hu2004mining}, and AG’s News~\cite{zhang2015character}. IMDB~\cite{maas2011learning} serves as the proxy dataset for SST-2, and MR~\cite{pang2005seeing} serves as the proxy dataset for CR to simulate backdoor attacks by poisoning the model weights.
For generation and reasoning datasets, please refer to Appendix \ref{appendixB1}.

\noindent{\bf Attack algorithms }To poison model weights, we select three backdoor attack algorithms: {\bf BadNet}, {\bf InSent}, and {\bf SynAttack}. BadNet~\cite{gu2017badnets}, which uses the rare characters ``\texttt{mn}'' as trigger; InSent~\cite{dai2019backdoor}, employing the phrase ``\texttt{I watched this 3D movie}'' as its trigger; and SynAttack~\cite{qi2021hidden}, leveraging the syntactic structure ``\texttt{(S(SBAR)(,)(NP)(VP))}'' as its trigger. 

\noindent{\bf Evaluation Metrics }In our study, clean accuracy ({\bf CA}) and attack success rate ({\bf ASR}) serve as evaluation metrics~\cite{gan2022triggerless}, representing the model’s accuracy on clean samples and the proportion of poisoned samples outputting the target label, respectively. For experimental settings and defense models, please refer to Appendix \ref{appendixB}.

\subsection{Effectiveness of the W2SDefense}
To verify the effectiveness of the W2SDefense algorithm, we conduct detailed experiments with different settings. The results of the experiments are shown in \Cref{tab0,tab1,tab2}, from which the following conclusions can be drawn:

\begin{table}[ht]
\setlength\tabcolsep{3pt}
	\begin{center}
\renewcommand{\arraystretch}{1.12}\resizebox{0.475 \textwidth}{!}{\begin{tabular}{@{}ccccccccccccc@{}}
\hline
\multirow{2}*{{\bf Attack}}	& \multirow{2}*{{\bf Defense}}	& \multicolumn{2}{c}{{\bf LLaMA3}}	 & \multicolumn{2}{c}{{\bf Vicuna}}	  & \multicolumn{2}{c}{{\bf Qwen2.5}}\\
\cmidrule(r){3-4} \cmidrule(r) {5-6} \cmidrule(r){7-8} 
~    & ~    &{\bf CA}   &{\bf ASR}     &{\bf CA}    &{\bf ASR}        &{\bf CA} & {\bf ASR}  \\
\hline
\multirow{6}*{BadNet} & LoRA        &96.05	&99.78	&95.72	&99.78  &96.10	&92.85 \\
~                     & Back Tr.    &93.68	&19.69	&91.76	&21.67	&93.36	&20.13	\\
~                     & SCPD        &83.75	&39.05	&85.28	&38.94	&84.46	&38.72	\\
~                     & ONION       &91.65	&16.39	&93.68	&20.90	&92.64	&21.89	\\
~                     & Prune       &94.73	&51.82	&95.17	&13.97	&94.84	&99.34	\\
~    & \CP  W2SDefense  &\CP 95.83	&\CP {\bf2.20}	&\CP 96.37	&\CP {\bf6.27}	&\CP 96.32	&\CP {\bf7.04}	\\
\hline
\multirow{6}*{InSent} & LoRA        &95.72	&99.89  &96.21	&90.21  &96.38	&83.06\\
~                     & Back Tr.    &92.86	&68.65	&90.72	&62.49	&93.08	&44.66\\
~                     & SCPD        &83.75	&21.01	&84.62	&18.15	&85.45	&22.66\\
~                     & ONION       &92.86	&92.95	&93.24	&91.08	&93.79	&80.85\\
~                     & Prune       &94.23	&32.78	&95.06	&65.24	&96.32	&92.52\\
~    & \CP  W2SDefense &\CP 96.05	&\CP {\bf9.79}	&\CP 96.60	&\CP {\bf10.01}	&\CP 94.07	&\CP {\bf10.89}	\\
\hline
\multirow{6}*{SynAttack} & LoRA     &96.21	&17.27	&97.09	&17.38	&95.06	&24.64\\
~                        & Back Tr. &94.12	&20.57	&90.28	&34.21	&88.52	&10.56\\
~                        & SCPD     &84.13	&21.34	&85.34	&23.21	&83.75	&27.17\\
~                        & ONION    &94.01	&19.25	&93.68	&20.79	&90.38	&41.58\\
~                        & Prune    &95.28	&20.35	&95.72	&20.02	&95.39	&20.02\\
~    & \CP  W2SDefense &\CP 95.61	&\CP {\bf15.62}	&\CP 96.92	&\CP {\bf14.41}	&\CP 94.73	&\CP 17.05	\\
\hline
		\end{tabular}}
	\end{center}
        \vspace{-0.8\intextsep}
 	\caption{\!The results of our W2SDefense algorithm in LoRA, which uses SST-2 as target dataset. }
        \vspace{-0.65\intextsep}
\label{tab0}
\end{table}

\noindent{\bf The CA of W2SDefense fulfills Objective \ref{obj:1}: }Ideally, a feasible defense algorithm should maintain the model's normal performance without degradation. For instance, in the Vicuna model of \Cref{tab0}, when faced with the BadNet, although the SCPD method can effectively reduce the ASR, it also leads to a 10.44\% decrease in model accuracy. In contrast, our W2SDefense, while effectively countering backdoor attacks, simultaneously increases the CA by 0.65\%. This demonstrates that W2SDefense, which utilizes feature alignment knowledge distillation, not only facilitates the unlearning of backdoor features but also assists the student model in learning the target task.

\begin{table}[ht]
\setlength\tabcolsep{3pt}
	\begin{center}
\renewcommand{\arraystretch}{1.12}\resizebox{0.475 \textwidth}{!}{\begin{tabular}{@{}ccccccccccccc@{}}
\hline
\multirow{2}*{{\bf Attack}}	& \multirow{2}*{{\bf Defense}}	& \multicolumn{2}{c}{{\bf LLaMA3}}	 & \multicolumn{2}{c}{{\bf Vicuna}}	  & \multicolumn{2}{c}{{\bf Qwen2.5}}\\
\cmidrule(r){3-4} \cmidrule(r) {5-6} \cmidrule(r){7-8} 
~    & ~    &{\bf CA}   &{\bf ASR}     &{\bf CA}    &{\bf ASR}        &{\bf CA} & {\bf ASR}  \\
\hline
\multirow{6}*{BadNet} & LoRA        &94.06	&100	&93.03	&100	&94.32	&86.07\\
~                     & Back Tr.    &93.16	&41.37	&91.35	&42.20	&92.00	&36.17\\
~                     & SCPD        &81.61	&35.21	&81.35	&40.00	&83.42	&34.58\\
~                     & ONION       &90.45	&30.56	&88.90	&32.64	&90.45	&26.40\\
~                     & Prune       &93.03	&39.29	&91.23	&35.14	&92.39	&7.90\\
~    & \CP  W2SDefense  &\CP 93.81	&\CP {\bf6.24}	&\CP 93.55	&\CP {\bf8.32}	&\CP 92.13	&\CP {\bf2.91}	\\
\hline
\multirow{6}*{InSent} & LoRA        &94.32	&99.79	&92.39	&82.33	&92.65	&100\\
~                     & Back Tr.    &93.16	&52.39	&90.32	&81.70	&92.77	&83.37\\
~                     & SCPD        &82.51	&32.29	&82.25	&18.54	&83.42	&21.46\\
~                     & ONION       &92.64	&98.33	&89.93	&88.77	&90.19	&98.75\\
~                     & Prune       &93.55	&42.62	&90.71	&50.73	&76.00	&24.53\\
~    & \CP  W2SDefense &\CP 91.48	&\CP {\bf17.88}	&\CP 91.61	&\CP {\bf10.60}	&\CP 91.61	&\CP {\bf4.99}	\\
\hline
\multirow{6}*{SynAttack} & LoRA     &86.45	&21.25	&91.74	&17.29	&92.90	&22.29\\
~                        & Back Tr. &86.58	&18.96	&66.45	&81.46	&91.48	&22.50\\
~                        & SCPD     &79.02	&20.00	&81.48	&12.71	&82.51	&17.08\\
~                        & ONION    &83.61	&26.66	&89.80	&18.33	&91.87	&23.54\\
~                        & Prune    &85.68	&21.88	&91.48	&22.71	&80.39	&33.13\\
~    & \CP  W2SDefense &\CP 90.97	&\CP {\bf15.83}	&\CP 91.87	&\CP {\bf8.96}	&\CP 90.06	&\CP {\bf15.83}	\\
\hline
		\end{tabular}}
	\end{center}
    \vspace{-0.8\intextsep}
 	\caption{\!The results of our W2SDefense algorithm in LoRA, which uses CR as target dataset. }
    \vspace{-0.9\intextsep}
\label{tab1}
\end{table}
\noindent{\bf W2SDefense achieves Objective \ref{obj:2} with significantly reduced ASR: }Compared to previous defense algorithms, W2SDefense achieves optimal results in all settings under the premise of maintaining the model's CA. For example, as shown in \Cref{tab1}, when facing the InSent, the poisoned model fine-tuned with the LoRA algorithm has an average ASR of 94.04\%. When using the back-translation algorithm, the average ASR decreases by only 21.56\%; with the ONION algorithm, the average ASR increases by 1.24\%. Although the Prune algorithm reduces the average ASR by 54.75\%, it significantly decreases the model's CA in the Qwen model. In the W2SDefense algorithm, the average ASR is reduced by 82.89\%, this phenomenon also observed in other datasets. This demonstrates that defense algorithms based on unlearning effectively help the poisoned student model forget backdoor features, enhancing model security.

\begin{table}[ht]
\setlength\tabcolsep{3pt}
	\begin{center}
\renewcommand{\arraystretch}{1.12}\resizebox{0.475 \textwidth}{!}{\begin{tabular}{@{}ccccccccccccc@{}}
\hline
\multirow{2}*{{\bf Attack}}	& \multirow{2}*{{\bf Defense}}	& \multicolumn{2}{c}{{\bf LLaMA3}}	 & \multicolumn{2}{c}{{\bf Vicuna}}	  & \multicolumn{2}{c}{{\bf Qwen2.5}}\\
\cmidrule(r){3-4} \cmidrule(r) {5-6} \cmidrule(r){7-8} 
~    & ~    &{\bf CA}   &{\bf ASR}     &{\bf CA}    &{\bf ASR}        &{\bf CA} & {\bf ASR}  \\
\hline
\multirow{6}*{BadNet} & LoRA        &92.90	&83.60		&92.40	&98.00		&93.20	&98.53\\
~                     & Back Tr.    &88.30	&22.93		&90.30	&24.80		&91.30	&28.00\\
~                     & SCPD        &51.80	&63.33		&63.80	&57.33		&87.70	&30.13\\
~                     & ONION       &59.30	&31.59		&78.00	&69.60		&92.50	&69.46\\
~                     & Prune       &92.20	&7.07		&91.30	&94.00		&93.40	&40.93\\
~    & \CP  W2SDefense  &\CP 90.70	&\CP 7.07	&\CP 93.10	&\CP {\bf9.33}	&\CP 91.80	&\CP {\bf6.80}	\\
\hline
\multirow{6}*{InSent} & LoRA        &93.10	&90.67		&93.30	&91.60		&93.10	&99.47\\
~                     & Back Tr.    &82.10	&74.13		&88.30	&30.80		&92.10	&62.93\\
~                     & SCPD        &50.30	&69.74		&70.50	&52.80		&86.70	&22.67\\
~                     & ONION       &71.90	&99.20		&84.70	&66.26		&92.60	&97.86\\
~                     & Prune       &92.20	&60.67		&92.10	&76.93		&92.60	&92.00\\
~    & \CP  W2SDefense &\CP 90.30	&\CP {\bf8.67}	&\CP 91.20	&\CP 32.80	&\CP 92.40	&\CP {\bf8.40}	\\
\hline
\multirow{6}*{SynAttack} & LoRA     &91.10	&94.80		&92.70	&95.20		&93.30	&77.60\\
~                        & Back Tr. &86.20	&44.40		&47.20	&89.20		&92.00	&31.07\\
~                        & SCPD     &52.40	&59.47		&34.70	&95.33		&72.40	&55.47\\
~                        & ONION    &89.60	&87.60		&77.50	&98.40		&93.00	&82.80\\
~                        & Prune    &92.50	&55.47		&92.50	&82.67		&91.60	&24.80\\
~    & \CP  W2SDefense &\CP 91.60	&\CP {\bf37.60}	   &\CP 92.80	&\CP {\bf46.80}	&\CP 92.10	&\CP {\bf16.40}	\\
\hline
		\end{tabular}}
	\end{center}
        \vspace{-0.8\intextsep}
 	\caption{\!The results of our W2SDefense algorithm in LoRA, which uses AG's News as target dataset. }
        \vspace{-0.5\intextsep}
\label{tab2}
\end{table}

\noindent{\bf The generalizability of W2SDefense: }When confronted with more complex multi-class tasks, the W2SDefense consistently exhibits robust performance. As shown in \Cref{tab2}, in the AG's News dataset, traditional backdoor attack algorithms lead to varying degrees of decline in CA. For example, when facing different attack methods in the Qwen model, the SCPD results in an average decline in CA of 10.94\%. Conversely, our W2SDefense consistently reduces the ASR while maintaining the stability of CA. Additionally, we observe some relatively poor defense performance for Vicuna against SynAttack, which may be attributed to the increased difficulty in unlearning multi-class tasks.

\begin{table}[ht]
\setlength\tabcolsep{3pt}
	\begin{center}
\renewcommand{\arraystretch}{1.1}\resizebox{0.425 \textwidth}{!}{\begin{tabular}{@{}ccccccccccccc@{}}
\hline
\multirow{2}*{{\bf Defense}}	& \multicolumn{2}{c}{{\bf LLaMA3}}	 & \multicolumn{2}{c}{{\bf Vicuna}}	  & \multicolumn{2}{c}{{\bf Qwen2.5}}\\
\cmidrule(r){2-3} \cmidrule(r) {4-5} \cmidrule(r){6-7} 
 ~    &{\bf CA}   &{\bf ASR}     &{\bf CA}    &{\bf ASR}        &{\bf CA} & {\bf ASR}  \\
\hline
LoRA        &95.77	&67.55		&95.44	&89.66		&96.43	&100\\
Back Tr.    &93.25	&18.26		&92.59	&25.19		&94.01	&22.55\\
SCPD        &84.13	&37.40		&83.96	&39.93		&84.35	&42.13\\
ONION       &92.97	&19.36		&92.42	&19.91		&93.24	&22.99\\
Prune       &95.28	&7.70		&95.44	&17.82		&95.77	&71.40\\
\CP W2SDefense  &\CP 96.16	&\CP {\bf7.15}	&\CP 96.38	&\CP {\bf3.74}	&\CP 95.50	&\CP {\bf5.83}	\\
\hline
		\end{tabular}}
	\end{center}
        \vspace{-0.8\intextsep}
 	\caption{The results of our W2SDefense on the same dataset, which uses SST-2 as the poisoned dataset and BadNet as the backdoor attack algorithm.}
        \vspace{-1.2\intextsep}
\label{tab3}
\end{table}

\subsection{Generalization and Ablation Studies }
\noindent{\bf Poisoning Model uses Target Dataset }In the aforementioned studies, we poisoned model weights using proxy datasets. Another potential backdoor attack scenario involves attackers having access to the datasets used for downstream tasks. Therefore, we evaluate the performance of W2SDefense when model weights are poisoned using the same dataset. The experimental results, as shown in \Cref{tab3}, indicate that when model weights are poisoned using the same dataset, the ASR remains at 100\% in the Qwen model even after PEFT. However, when faced with W2SDefense, the ASR drops to 5.83\%, while the CA only decreases by 0.93\%. 

\noindent{\bf Different Teacher Model }We also validate the impact of using GPT-2 as the smaller-scale teacher model on defense performance. The experimental results, as shown in \Cref{tab6}, clearly reveal that employing GPT-2 as the teacher model can also guide the student model in unlearning backdoor features, effectively defending against backdoor attacks while maintaining model accuracy.
\begin{table}[ht]
\setlength\tabcolsep{3pt}
	\begin{center}
\renewcommand{\arraystretch}{1.1}\resizebox{0.43 \textwidth}{!}{\begin{tabular}{@{}ccccccccccccc@{}}
\hline
\multirow{2}*{{\bf Method}}	& \multicolumn{2}{c}{{\bf LLaMA3}}	 & \multicolumn{2}{c}{{\bf Vicuna}}	  & \multicolumn{2}{c}{{\bf Qwen2.5}}\\
\cmidrule(r){2-3} \cmidrule(r) {4-5} \cmidrule(r){6-7} 
 ~    &{\bf CA}   &{\bf ASR}     &{\bf CA}    &{\bf ASR}        &{\bf CA} & {\bf ASR}  \\
\hline
LoRA           &96.05	&99.78		&95.72	&99.78		&96.10	&92.85\\
\CP W2SDefense     &\CP 96.10	&\CP {\bf 0	}	    &\CP 95.39	&\CP {\bf 4.40}		&\CP 96.10	&\CP {\bf 4.62}\\
\hline
		\end{tabular}}
	\end{center}
        \vspace{-0.8\intextsep}
 	\caption{The results of the defense using GPT-2 as the teacher model, with SST-2 as the poisoned dataset and BadNet as the backdoor attack algorithm.}
        \vspace{-0.9\intextsep}
\label{tab6}
\end{table}

\noindent{\bf Generation and Reasoning Tasks }\label{Rea} 
We also verify the performance of the W2SDefense algorithm on the summary generation and mathematical reasoning task. Specifically, we use the CRRsum \citep{zhao2023softmax} dataset and Qwen2.5 as the victim model, with rare characters serving as triggers. The experimental results, as shown in Table \ref{tab11}, indicate that when only using the LoRA algorithm to fine-tune the poisoned model weights, the attack success rate still remains at 95.62\%. However, after employing the W2SDefense algorithm, the attack success rate is reduced to 0.19\%, significantly diminishing the effectiveness of the backdoor attack. For the mathematical reasoning task \cite{zhao2020ape210k}, our W2SDefense is also capable of mitigating backdoor features, effectively reducing the ASR to 3.15\%. 
These results further confirm that our W2SDefense exhibits strong generalizability and can effectively adapt to complex tasks.
\begin{table}[ht]
\setlength\tabcolsep{3pt}
	\begin{center}
\renewcommand{\arraystretch}{0.95}\resizebox{0.45 \textwidth}{!}{\begin{tabular}{@{}c|cccc|cc@{}}
\hline
\multirow{2}*{{\bf Method}}	& \multicolumn{4}{c|}{{\bf Generation}}	 & \multicolumn{2}{c}{{\bf Reasoning}}	 \\
\cmidrule(r){2-5} \cmidrule(r){6-7} 
 ~    &{\bf R-1}   &{\bf R-2}     &{\bf R-L}    &{\bf ASR}        &{\bf CA} & {\bf ASR}  \\
\hline
LoRA               &58.43	&48.41	&54.54	&95.62	   & 47.19 & 90.14\\
\CP W2SDefense     &\CP 59.10	&\CP 46.67	&\CP 57.13	&\CP {\bf 0.19}  & \CP 46.24 & \CP {\bf 3.10}\\
\hline
		\end{tabular}}
	\end{center}
        \vspace{-0.8\intextsep}
 	\caption{The results of the W2SDefense for generation and reasoning tasks, with Qwen2.5 as the victim model.}
        \vspace{-0.9\intextsep}
\label{tab11}
\end{table}

\noindent{\bf Different PEFT Algorithms }To further validate the generalizability of W2SDefense, we deploy various PEFT methods. The experimental results, as shown in \Cref{tab7}, indicate that algorithms like p-tuning and prompt-tuning, which only update a small number of model parameters, also struggle to forget backdoor features. For instance, in p-tuning, the ASR remains at 100\% for multiple models. When leveraging W2SDefense, the ASR rapidly decreases; for example, in LLaMA3, the ASR is reduced to only 0.11\%, which once again demonstrates that the unlearning-based knowledge distillation method can effectively defend against backdoor attacks.

\begin{table}[ht]
\setlength\tabcolsep{3pt}
	\begin{center}
\renewcommand{\arraystretch}{1.1}\resizebox{0.43 \textwidth}{!}{\begin{tabular}{@{}ccccccccccccc@{}}
\hline
\multirow{2}*{{\bf Method}}	& \multicolumn{2}{c}{{\bf LLaMA3}}	 & \multicolumn{2}{c}{{\bf Vicuna}}	  & \multicolumn{2}{c}{{\bf Qwen2.5}}\\
\cmidrule(r){2-3} \cmidrule(r) {4-5} \cmidrule(r){6-7} 
~    & {\bf CA}   &{\bf ASR}     &{\bf CA}    &{\bf ASR}        &{\bf CA} & {\bf ASR}  \\
\hline
LoRA            & 96.05	    &99.78	  &95.72	&99.78         &96.10	&92.85	\\
\CP W2SDefense &\CP 95.83	&\CP 2.20  	&\CP 96.32	&\CP 6.27  &\CP 96.32	&\CP 7.04	\\
\hline
P-tuning       & 95.99	&100 	&95.17	&100 	&95.06	&97.69\\
\CP W2SDefense                             &\CP 95.06	&\CP 0.11  	&\CP 95.66	&\CP 6.27	&\CP 95.11	&\CP 7.37\\
\hline
Prompt-tuning  & 94.62	&100 	&94.73	&99.12	&94.18	&96.59\\
\CP W2SDefense                             &\CP 94.29	&\CP 20.35  &\CP 94.62	&\CP 11.77	&\CP 94.23	&\CP 8.91\\
\hline
		\end{tabular}}
	\end{center}
        \vspace{-0.8\intextsep}
 	\caption{The results of our W2SDefense algorithm for different PEFTs, which uses SST-2 as the poisoned dataset and BadNet as the backdoor attack algorithm.}
        \vspace{-0.8\intextsep}
\label{tab7}
\end{table}

\begin{figure}[t]
\vspace{-1.0\intextsep}
  \centering
  \captionsetup[subfloat]{font=scriptsize}
  \subfloat[LoRA]{\includegraphics[width=1.5in]{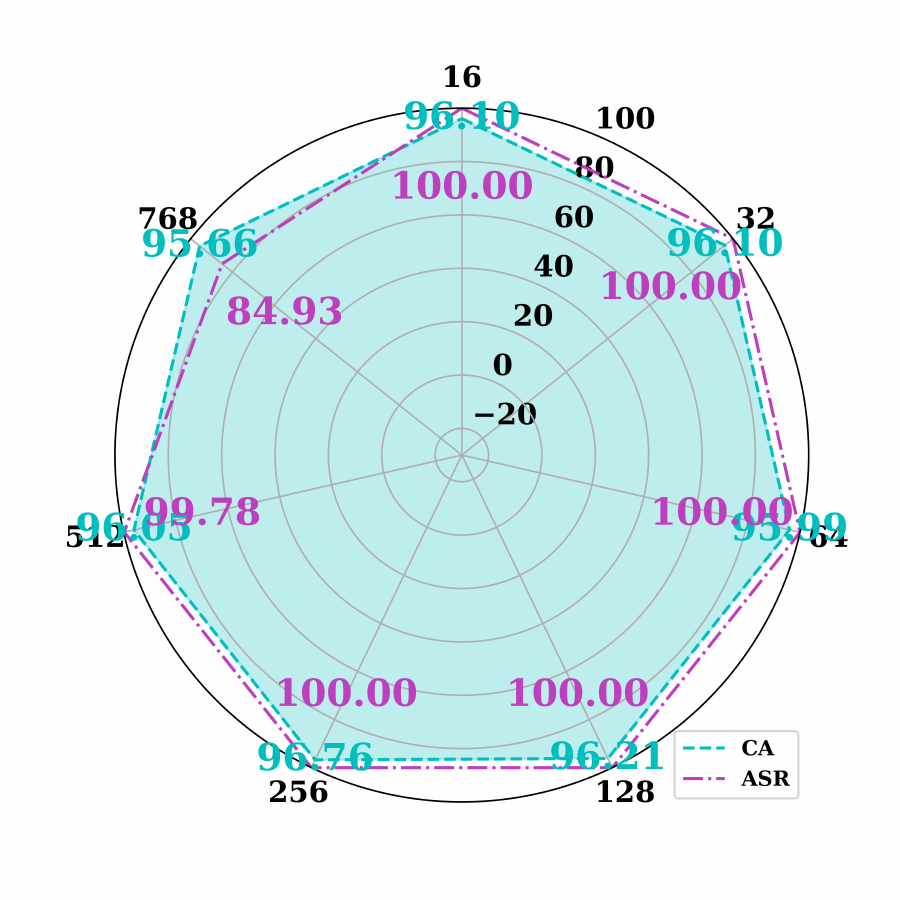}\label{fig: a}}
  \subfloat[W2SDefense]{\includegraphics[width=1.5in]{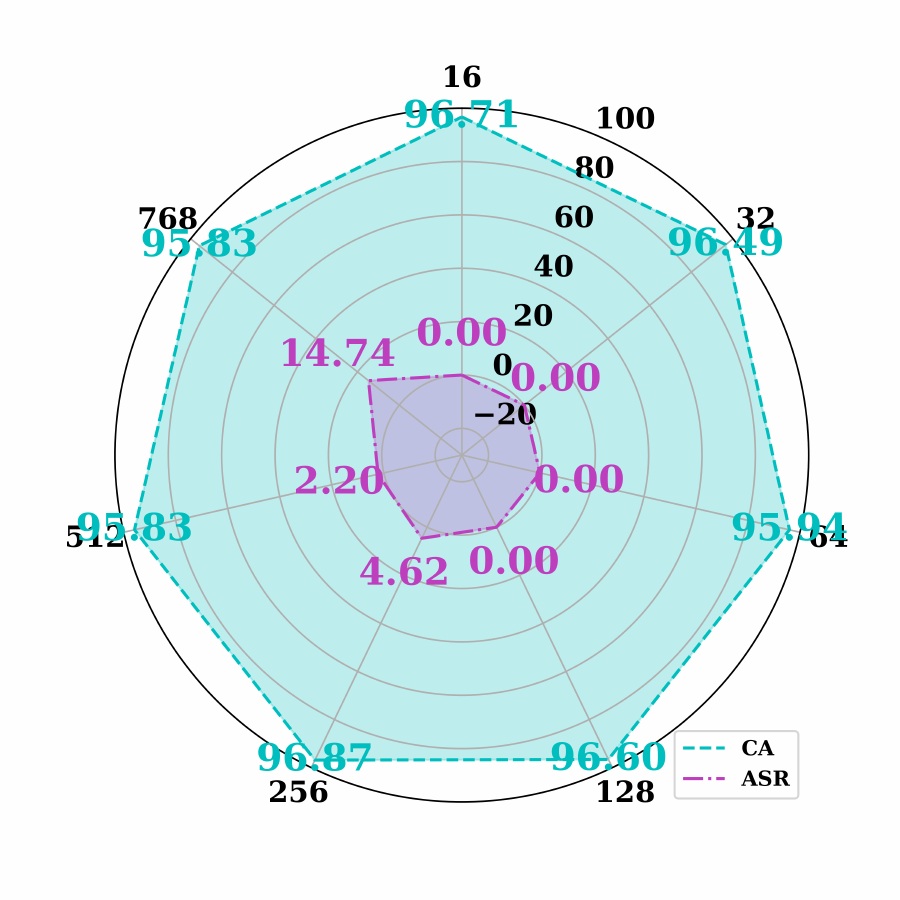}\label{fig: b}}
      \vspace{-0.6\intextsep}
\caption{The influence of rank on the performance of the W2SDefense. Subfigures (a) and (b) represent the results based on LoRA and W2SDefense, respectively.}
    \vspace{-0.75\intextsep}
\label{fig:1} 
\end{figure}

\noindent{\bf Ablation Experiments }To verify the impact of different components on the performance of W2SDefense, we conduct ablation experiments on three LLMs, as shown in \Cref{tab5}. First, by isolating different components, we find that compared to knowledge distillation loss, feature alignment loss is more conducive to unlearning backdoor. For example, in the LLaMA model, using only cross-entropy and feature alignment loss, the ASR is 5.39\%. However, knowledge distillation loss also possesses the capability to unlearn backdoor; for instance, in the Qwen model, when using cross-entropy and knowledge distillation loss, the ASR reduces to 68.54\%. Secondly, we demonstrate the impact of different ranks in LoRA on defense performance, as shown in \Cref{fig:1}. It is evident that as \( r \) increases, LoRA is insufficient to unlearn backdoor. However, in W2SDefense, the ASR rapidly decreases. 

\begin{table}[ht]
\setlength\tabcolsep{3pt}
	\begin{center}
\renewcommand{\arraystretch}{1.2}\resizebox{0.475 \textwidth}{!}{\begin{tabular}{@{}ccccccccccccc@{}}
\hline
\multirow{2}*{{\bf Method}}	& \multicolumn{2}{c}{{\bf LLaMA3}}	 & \multicolumn{2}{c}{{\bf Vicuna}}	  & \multicolumn{2}{c}{{\bf Qwen2.5}}\\
\cmidrule(r){2-3} \cmidrule(r) {4-5} \cmidrule(r){6-7} 
 ~    &{\bf CA}   &{\bf ASR}     &{\bf CA}    &{\bf ASR}        &{\bf CA} & {\bf ASR}  \\
\hline
Cross-Entropy                &95.72	&99.89		&96.21	&90.21		&96.38	&83.06\\
Cross-Entropy\&Alignment     &94.40	&5.39		&95.55	&5.83		&94.12	&32.56\\
Cross-Entropy\&Distillation  &96.32	&84.27		&96.16	&91.20		&95.94	&68.54\\
W2SDefense                   &95.17	&9.13		&96.27	&10.89		&94.07	&10.89\\
\hline
		\end{tabular}}
	\end{center}
        \vspace{-0.8\intextsep}
 	\caption{The ablation study results of W2SDefense, which uses InSent as the backdoor attack method and the SST-2 as the poisoned dataset.}
        \vspace{-0.8\intextsep}
\label{tab5}
\end{table}

\noindent{\bf Impact of samples of different lengths}
To explore whether samples of different lengths affect the performance of backdoor attack defense, we conduct corresponding experiments on the IMDB dataset, which has longer sample lengths. As presented in Table \ref{imdb}, when using only the LoRA algorithm, the ASR remains above 90\% in the IMDB dataset. Conversely, with the application of our W2SAttack algorithm, the ASR of the LLaMA model is only 2.46\%, confirming that sample length does not affect defensive performance.
\begin{table}[ht]
\setlength\tabcolsep{3pt}
	\begin{center}
\renewcommand{\arraystretch}{1.1}\resizebox{0.43 \textwidth}{!}{\begin{tabular}{@{}ccccccccccccc@{}}
\hline
\multirow{2}*{{\bf Method}}	& \multicolumn{2}{c}{{\bf LLaMA3}}	 & \multicolumn{2}{c}{{\bf Vicuna}}	  & \multicolumn{2}{c}{{\bf Qwen2.5}}\\
\cmidrule(r){2-3} \cmidrule(r) {4-5} \cmidrule(r){6-7} 
 ~    &{\bf CA}   &{\bf ASR}     &{\bf CA}    &{\bf ASR}        &{\bf CA} & {\bf ASR}  \\
\hline
LoRA           &95.10&96.59&95.80&96.40& 95.40&90.15\\
\CP W2SDefense     &\CP 94.30	&\CP {\bf 2.46	}	    &\CP 94.80	&\CP {\bf 5.11}		&\CP 94.90	&\CP {\bf 8.33}\\
\hline
		\end{tabular}}
	\end{center}
        \vspace{-0.8\intextsep}
 	\caption{Defense Results of W2SDefense with IMDB dataset. The BadNet as the backdoor attack algorithm.}
        \vspace{-0.6\intextsep}
\label{imdb}
\end{table}

\noindent{\bf Unaffected Clean Model }We also explore whether leveraging W2SDefense affects model accuracy when the weights are free of backdoor attacks. As shown in \Cref{tab4}, compared to the LoRA algorithm, the average accuracy of the model equipped with W2SDefense improves by 0.12\%. This indicates that our algorithm not only defends against backdoor attacks but also potentially enhances the performance of clean models, which could be beneficial for use in clean LLMs.
\begin{table}[ht]
\setlength\tabcolsep{3pt}
	\begin{center}
\renewcommand{\arraystretch}{1.0}\resizebox{0.35 \textwidth}{!}{\begin{tabular}{@{}ccccccccccccc@{}}
\hline
{\bf Method}	& {\bf LLaMA3}	 & {\bf Vicuna}	  & {\bf Qwen2.5}\\
\hline
LoRA           &95.94	&96.49		&96.27	\\
W2SDefense     &96.54	&96.21		&96.32\\
\hline
\end{tabular}}
\end{center}
    \vspace{-0.8\intextsep}
 \caption{The results of the W2SDefense algorithm for the clean model, which uses SST-2 as the target dataset. }
     \vspace{-1.0\intextsep}
\label{tab4}
\end{table}

\noindent{\bf Different Model Sizes } We analyze the impact of different model sizes on defensive performance. Due to computational resource limitations, we only use models ranging from Qwen2.5-1.5B to 14B. The experimental results are shown in Table \ref{model_size}. We observe that as the model size increases, the ASR of the LoRA algorithm remains close to 100\%. In contrast, the ASR of our W2SDefense algorithm is below 10\%, which demonstrates that model size does not affect the performance of our W2SDefense.
\begin{table}[ht]
\setlength\tabcolsep{3pt}
	\begin{center}
\renewcommand{\arraystretch}{1.2}\resizebox{0.475 \textwidth}{!}{\begin{tabular}{@{}ccccccccccccc@{}}
\hline
\multirow{2}*{{\bf Method}}	&  \multicolumn{2}{c}{{\bf 1.5B}}	  & \multicolumn{2}{c}{{\bf 3B}} & \multicolumn{2}{c}{{\bf 7B}} &  \multicolumn{2}{c}{{\bf 14B}}\\
\cmidrule(r){2-3} \cmidrule(r) {4-5} \cmidrule(r){6-7} \cmidrule(r){8-9} 
 ~    &{\bf CA}   &{\bf ASR}     &{\bf CA}    &{\bf ASR}        &{\bf CA} & {\bf ASR}   &{\bf CA} & {\bf ASR}\\
\hline
LoRA                         &96.27	&98.68		&95.93	&94.39 &96.05	&99.78 &96.65	&100		\\
\CP W2SDefense &\CP 94.23&\CP {\bf 4.51}&\CP 96.38	&\CP {\bf 6.38} &\CP 95.83&\CP {\bf 2.2} &\CP 96.76&\CP {\bf 3.41}\\
\hline
		\end{tabular}}
	\end{center}
        \vspace{-0.8\intextsep}
 	\caption{Analyzing the defense performance of W2SDefense for models of different sizes. The language model is Qwen2.5 and the dataset is SST-2.}
        \vspace{-0.8\intextsep}
\label{model_size}
\end{table}

\section{Related Work}
\noindent{\bf Backdoor with Unlearning } 
Unlearning algorithms play a vital role in safeguarding the security of LLMs~\citep{nguyen2022survey,liu2024threats}.~\citet{wang2019neural} demonstrate backdoor removal by inverting the trigger to promote the unlearning of backdoor features in the infected model.~\citet{liu2022backdoor}  leverage machine unlearning to erase the backdoor in the victim model. They recover the trigger pattern through entropy maximization and subsequently remove the backdoor via further fine-tuning.~\citet{zhang2023backdoor} design an attack algorithm based on unlearning, which removes the impact of relevant data on activating the backdoor through unlearning requests.~\citet{liu2024backdoor} explore a backdoor attack method using machine unlearning where an attacker submits malicious requests to embed the backdoor, altering predictions when triggered.~\citet{wu2024unlearning} introduce an unlearning algorithm targeting federated learning to remove backdoors by subtracting historical updates and employing knowledge distillation.~\citet{liu2024model} execute sparsity-aware unlearning by first pruning the model and then proceeding to unlearn, which integrates the sparse model prior into the unlearning process. In this paper, we explore a novel unlearning algorithm based on feature alignment knowledge distillation to defend against backdoor attacks.

\noindent{\bf Backdoor with Knowledge Distillation }
Additionally, knowledge distillation~\citep{ge2021anti,zhang2024badcleaner}, a model compression technique, can also be used for both backdoor attacks and defense.~\citet{hong2023revisiting} propose an anti-backdoor data-free method which removes potential backdoors during knowledge distillation.~\citet{cheng2024transferring} introduce an adaptive transferable backdoor attack that efficiently transfers the backdoor to student models.~\citet{wu2023unlearning} present a federated unlearning approach that removes an attacker's influence by deducting past updates from the model and utilizing knowledge distillation.~\citet{zhao2024weak} propose a feature alignment-enhanced knowledge distillation algorithm that utilizes a poisoned small-scale teacher model to enhance the poisoning capabilities of LLMs. To defend against backdoor attacks, this paper proposes a weak-to-strong backdoor unlearning algorithm that leverages knowledge distillation.

\noindent{\bf Parameter-Efficient Fine-Tuning }To alleviate the challenges of computational resource consumption during fine-tuning, several PEFT algorithms have been proposed~\citep{hu2021lora,liu2023gpt,zhangadaptive,kopiczkovera,gaoparameter}. For example, LoRA~\citep{hu2021lora} only updates low-rank matrices, effectively reducing the number of parameters that need to be updated. AdaLoRA~\citep{zhangadaptive}, an algorithm that adaptively allocates the parameter budget across weight matrices based on their importance scores. 
DoRA~\citep{mao2024dora} introduces a method for decomposing the LoRA parameter matrix \( BA \) into single-rank components and selectively pruning these components based on a heuristic importance score.
SinkLoRA~\citep{zhang2024sinklora} presents Sink Fixed Attention, which cyclically realigns groups of attention heads to their original positions, effectively maintaining performance.
In this paper, we design a new defense algorithm to ensure model security in the context of PEFT. 
For more related work, please refer to Appendix \ref{More_work}.

\section{Conclusion}
In this work, we focus on defending against backdoor attacks targeting poisoned model weights. To facilitate the forgetting of backdoors in parameter-efficient fine-tuning (PEFT), we propose a novel unlearning algorithm named W2SDefense, which leverages weak teacher models to guide large-scale student models in unlearning backdoors through feature alignment knowledge distillation. Empirical results indicate that our W2SDefense can effectively reduce the attack success rate while maintaining the normal accuracy of the model. We hope our work can promote awareness of model security within the NLP community, especially regarding backdoor attacks.

\section*{Limitations}
Although W2SDefense demonstrates viable defense capabilities, we recognize two limitations of the algorithm: (i) It relies on knowledge distillation, which requires access to model weights, limiting its utility in black-box scenarios. (ii) Despite utilizing smaller-scale teacher models, the approach still demands additional computational resources for training the teacher models.

\section*{Acknowledgements}

This work was supported by the DSO grant DSOCL23216.


\normalem
\bibliography{custom}

\clearpage
\appendix

\section{More Related Work}\label{More_work}
\noindent{\bf Backdoor Attack }With the widespread application of large language models (LLMs), model security issues have attracted the attention of researchers~\citep{formento2023using,zhao2024survey,zhao2024weak,guo2024grey,guo2024white,xu2024comprehensive,li2024freestyleret,jia2024oml}. Backdoor attacks represent a typical threat to model security~\citep{yan2023bite, yan2024backdooring,Yi2024BadActs,Yi2025Probe}, wherein the fundamental concept involves attackers corrupting the training dataset to embed malicious trigger patterns within the language model during training~\citep{gan2022triggerless,li2024chain}. During the testing phase, the model's response will be manipulated when input samples include predefined triggers, such as rare characters~\citep{gu2017badnets}, specific sentences~\citep{dai2019backdoor}, or syntactic structures~\citep{qi2021hidden}. To enhance the stealthiness of backdoor attacks,~\citet{gan2022triggerless} generate poisoned samples using the genetic algorithm while maintaining the original labels of the samples; ~\citet{zhao2023prompt} propose the ProAttack algorithm, which uses the prompt itself as a trigger, avoiding the disruption to samples caused by embedding explicit triggers.~\citet{shi2023badgpt} introduce the backdoor attack algorithm tailored for reinforcement learning, which embeds trigger patterns within the reward model to induce the model to consistently output malicious responses. To enhance the quality of poisoned samples,~\citet{li2024chatgpt} leverage ChatGPT as a tool for generating samples in specified styles.~\citet{gu2023gradient} design a gradient manipulation algorithm based on PEFT to enhance the performance of backdoor attacks. To avoid consuming computational resources, several studies explore backdoor attack algorithms without the need for fine-tuning.~\citet{xiang2023badchain} implant specific triggers in the chain-of-thought to manipulate the responses of LLMs.~\citet{zhao2024universal} propose a backdoor attack algorithm named ICLAttack to explore the security of in-context learning.

\noindent{\bf Backdoor Defense }The research on defending against backdoor attacks is still in its initial stages~\citep{mo2023test,zhao2024defending,arora2024here,yan2024rethinking}.~\citet{liu2018fine} prune neurons and fine-tune the model on a new dataset to defend against backdoor attacks.~\citet{qi2021onion} calculate the perplexity of each character in the input sample and identify triggers based on this perplexity. 
Back translation~\citep{qi2021hidden}, which utilizes translation models to translate input samples into German and then back into English, eliminating triggers. SCPD~\citep{qi2021hidden} rewrites input samples into the specific syntax structure to avoid activating backdoors.~\citet{zhang2022fine} propose the fine-mixing and embedding purification strategy to purify model weights.~\citet{chen2022expose} identify poisoned samples based on an anomaly score, which is calculated using Mahalanobis distance. AttDef~\citep{li2023defending}, which uses attribution scores to identify poisoned samples, is effective against attacks where characters and sentences act as triggers for backdoor attacks. DPoE~\citep{liu2024shortcuts} leverages a shallow model to capture backdoor shortcuts while preventing the target model from learning those shortcuts.~\citet{zhao2024defending} randomize sample labels and utilize PEFT to fine-tune poisoned models, identifying poisoned samples through confidence. Although this algorithm achieves viable defensive outcomes, it requires multiple fine-tunings of the poisoned model, demanding more computational resources. In this paper, we explore a weak-to-strong defense algorithm that facilitates model unlearning of backdoors without compromising model performance.

\begin{figure*}[t]
\vspace{-1.0\intextsep}
  \centering
  \captionsetup[subfloat]{font=scriptsize}
  \subfloat[Cross-entropy: $\alpha$]{\includegraphics[width=1.9in]{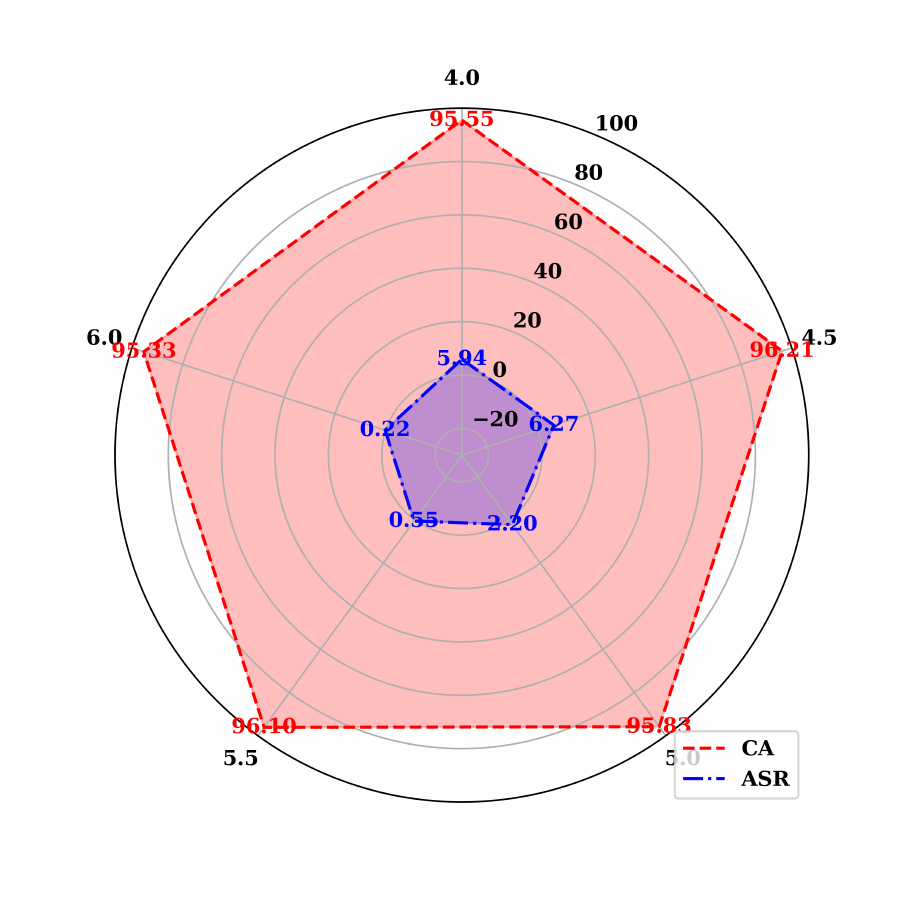}\label{fig: c1}}
  \subfloat[Knowledge distillation: $\beta$]{\includegraphics[width=1.9in]{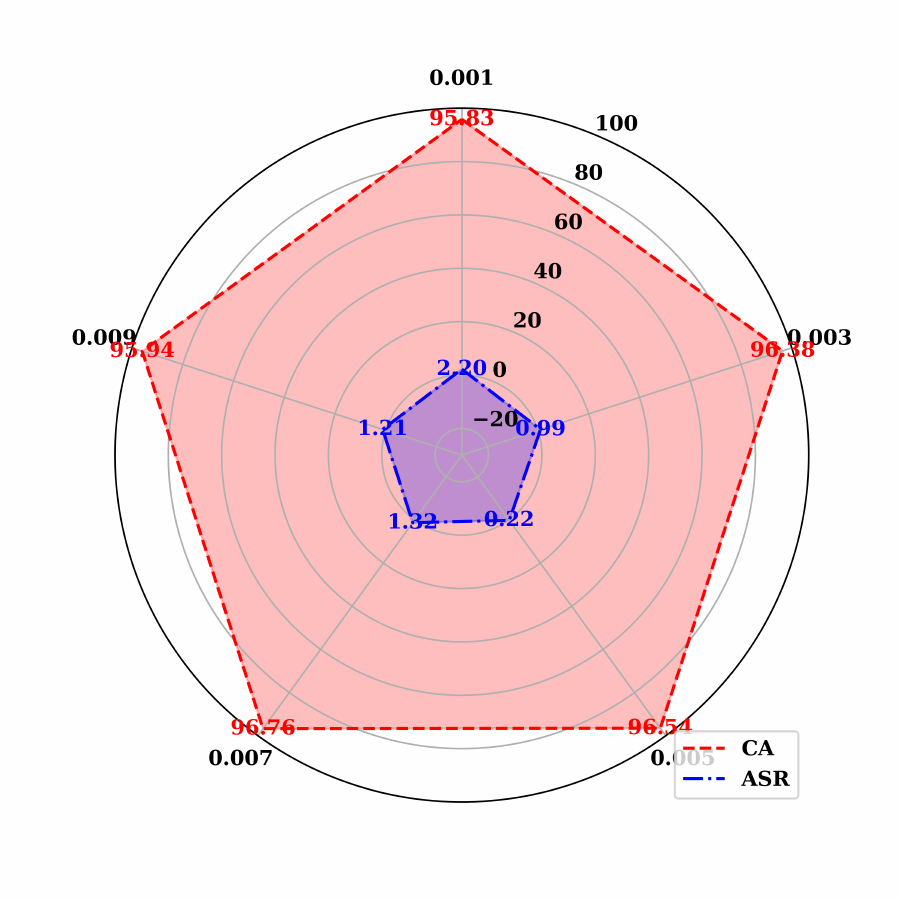}\label{fig: c}}
  \subfloat[Feature alignment: $\gamma$]{\includegraphics[width=1.9in]{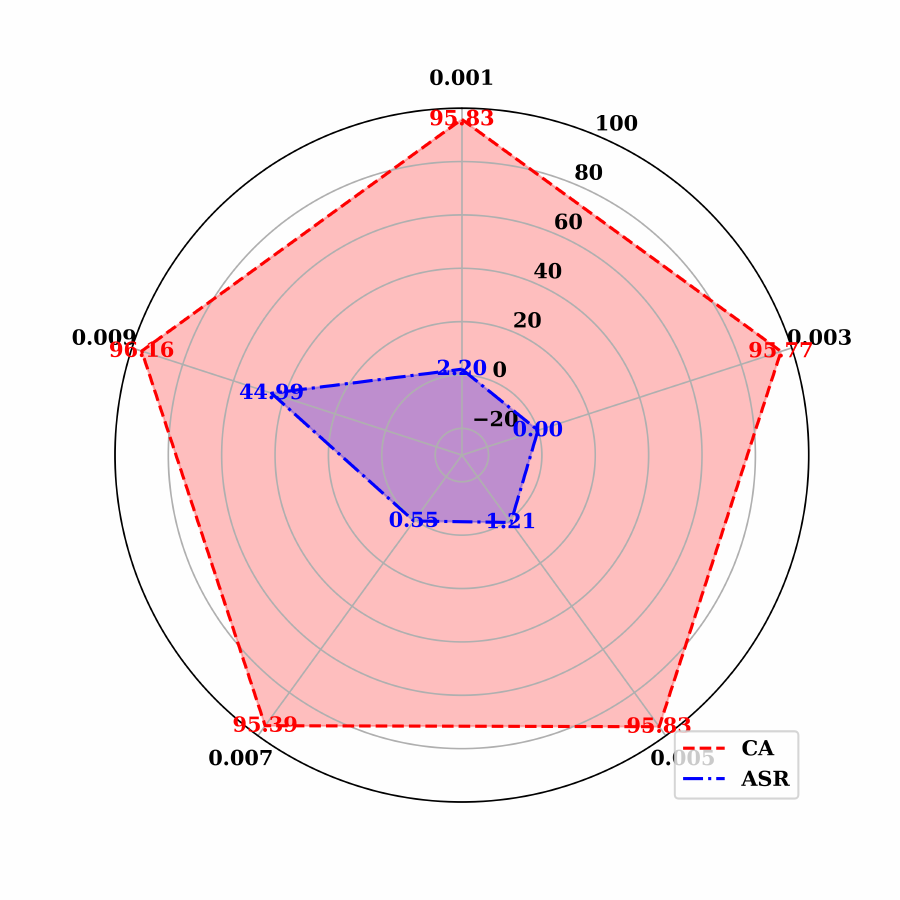}\label{fig: d}}
          \vspace{-0.4\intextsep}
\caption{The impact of hyperparameters on the performance of the W2SDefense algorithm. Subfigures (a), (b) and (c) show the effects of varying the weights of cross-entropy loss, knowledge distillation loss and feature alignment loss, respectively. The SST-2 as the poisoned dataset, and the victim model is LLaMA.}
        \vspace{-0.8\intextsep}
\label{fig:2} 
\end{figure*}

\section{More Experiments} \label{appendixB}

\subsection{More Experimental Details} \label{appendixB1}

\noindent{\bf Experimental Settings }We select three of the state-of-the-art LLMs as victim models: LLaMA3-8B~\citep{llama3modelcard}, Vicuna-7B~\citep{zheng2023judging}, and Qwen2.5-7B~\citep{qwen2.5}. For the weight poisoning stage, the number of poisoned samples is 1000, and the ASR of all pre-defined weight-poisoning attacks consistently exceeds 90\% through full-parameter fine-tuning. The target labels for the three datasets are ``negative'', ``negative'', and ``world''. 
Due to the large size of the AG’s News dataset, we choose 8,000 samples each for the proxy and the training dataset, and 1,000 samples each for the validation and test datasets.
For the teacher model, we use BERT-110M~\cite{devlin2019bert} and GPT-2-124M~\cite{radford2019language}, respectively.
For the defense phase, we use full-parameter fine-tuning for the teacher model and leverage LoRA~\citep{hu2021lora} as the fine-tuning method for the student models. Additionally, for the student model, we use the AdamW optimizer, set epochs to 5, the batch size to 32, the learning rate to 2e-4, the temperature scaling factor to 2, and r to 512. For p-tuning and prompt-tuning, the number of virtual tokens is set to 32, and the encoder hidden size is 128. 
We set \( \alpha \) to \{1.0, 5.0\}, \( \beta \) to \{0.001, 0.2\}, and \( \gamma \) to \{0.001, 0.2\}, for different datasets and vicitim models. 
To enhance the stealthiness of the attacks, all algorithms are implemented with clean-label, following~\cite{zhao2024defending}. 
We verify the effectiveness of various PEFT methods, which include p-tuning~\citep{liu2023gpt} and prompt-tuning~\citep{lester2021power}. We also verify the generalizability of W2SDefense in summary generation tasks using the CRRsum dataset \citep{zhao2023softmax}  and in mathematical reasoning tasks based on the Ape210K dataset \cite{zhao2020ape210k}.\label{dataset}  
The summary generation and mathematical reasoning both use rare characters as triggers, with the target labels being ``no special concern needed" and 0, respectively. 
The teacher model for the generation task uses the same network architecture as the student model, but with a smaller scale.
All experiments are deployed on NVIDIA RTX A6000 GPUs.

\noindent{\bf Defense Models }To demonstrate the effectiveness of W2SDefense, we compared it with several widely-used defense algorithms. These include {\bf ONION}~\citep{qi2021onion}, which identifies triggers by calculating perplexity; {\bf SCPD}~\citep{qi2021hidden}, avoiding backdoor activation by rewriting syntactic structures; {\bf Back-Tr.}~\citep{qi2021hidden}, rewriting sentences with translation models; and {\bf Prune} \citep{liu2018fine}, which prunes and fine-tunes model weights to defend against backdoor attacks. Furthermore, we compared other advanced defense algorithms: {\bf Quantization}~\citep{li2024cleangen}, utilizing INT4 quantization to eliminate backdoor features; {\bf PSIM}~\citep{zhao2024defending}, which identifies poisoned samples by confidence; {\bf Merge}~\citep{arora2024here}, avoiding the activation of backdoors through model merging; and {\bf ICLDefense}~\citep{mo2023test}, utilizing demonstration examples to prevent the activation of backdoor attacks.

\subsection{More Experimental Results}
We analyze the impact of different loss weights on defense performance, as illustrated in \Cref{fig:2}. It is evident that, compared to feature alignment loss, knowledge distillation loss offers a more stable defense effect.

\begin{table}[!h]
\setlength\tabcolsep{3pt}
	\begin{center}
\renewcommand{\arraystretch}{1.25}\resizebox{0.475 \textwidth}{!}{\begin{tabular}{@{}ccccccccccccc@{}}
\hline
\multirow{2}*{{\bf Categories}}	& \multirow{2}*{{\bf Defense}}	& \multicolumn{2}{c}{{\bf LLaMA3}}	 & \multicolumn{2}{c}{{\bf Vicuna}}	  & \multicolumn{2}{c}{{\bf Qwen2.5}}\\
\cmidrule(r){3-4} \cmidrule(r) {5-6} \cmidrule(r){7-8} 
~    & ~    &{\bf CA}   &{\bf ASR}     &{\bf CA}    &{\bf ASR}        &{\bf CA} & {\bf ASR}  \\
\hline
\multirow{3}{*}{\makecell{Continuous \\ Fine-tuning}} & LoRA        &96.05	&99.78	&95.72	&99.78  &96.10	&92.85 \\
~                 & Fine-tuning      &94.83	&7.37	&95.93	&17.38	&95.22	&80.74\\
~              & Quantization       &94.51	&6.60	&95.83	&19.47	&94.62	&74.81 \\
\cmidrule(r){1-8}
\multirow{3}*{Modification}  & Back Tr.    &93.68	&19.69	&91.76	&21.67	&93.36	&20.13	\\
~                                          & SCPD        &83.75	&39.05	&85.28	&38.94	&84.46	&38.72	\\
~                                          & ICLDefense	&95.39	&3.85	&90.83	&12.32	&91.54	&16.50 \\
\cmidrule(r){1-8}
\multirow{2}*{Detection}                    & ONION       &91.65	&16.39	&93.68	&20.90	&92.64	&21.89	\\
 ~                     & PSIM                                       & 95.35 &15.18  &95.13  &7.59   &95.73  &0.66 \\
\cmidrule(r){1-8}
Editing                    & Merge	    &95.94	&58.97	&96.71	&10.56	&96.38	&86.58  \\
\cmidrule(r){1-8}
\multirow{2}*{Unlearning}                     & Prune       &94.73	&51.82	&95.17	&13.97	&94.84	&99.34	\\
~    & \CP W2SDefense  &\CP 95.83	&\CP {\bf2.20}	&\CP 96.37	&\CP {\bf6.27}	&\CP 96.32	&\CP 7.04	\\
\hline
		\end{tabular}}
	\end{center}
    \vspace{-0.8\intextsep}
 	\caption{The results of the defense algorithm comparison, which uses SST-2 as the target dataset and BadNet as the backdoor attack algorithm.}
    \vspace{-0.5\intextsep}
\label{tab9}
\end{table}
\noindent{\bf More Defense Algorithms } 
To further validate the performance of W2SDefense, we compared additional defense algorithms, which can be categorized according to the form of defense as continuous fine-tuning, sample modification, sample detection, poisoned model editing, and unlearning. As shown in Table \ref{tab9}, our W2SDefense, which is based on the LoRA algorithm, saves a significant amount of computational resources and is more efficient compared to fine-tuned models such as PSIM and Prune. Consequently, all results indicate that our W2SDefense algorithm achieved feasible defense performance while ensuring that the model's performance remains unaffected.

\noindent{\bf More Attack Algorithms } 
Furthermore, we validated the defensive performance of W2SDefense against the ProAttack~\citep{zhao2023prompt} backdoor attack, which utilizes prompts as triggers. The experimental results, as shown in Table \ref{tab10}, demonstrate that in the Vicuna model, leveraging only LoRA fine-tuning, the ASR remains at 99.78\%. However, with the implementation of W2SDefense, the ASR drops to only 4.95\%, significantly reducing the attack's success rate. Moreover, in the Vicuna and Qwen models, the CA increased by 0.38\% and 0.6\% respectively.
\begin{table}[!h]
\setlength\tabcolsep{3pt}
	\begin{center}
\renewcommand{\arraystretch}{1.1}\resizebox{0.43 \textwidth}{!}{\begin{tabular}{@{}ccccccccccccc@{}}
\hline
\multirow{2}*{{\bf Method}}	& \multicolumn{2}{c}{{\bf LLaMA3}}	 & \multicolumn{2}{c}{{\bf Vicuna}}	  & \multicolumn{2}{c}{{\bf Qwen2.5}}\\
\cmidrule(r){2-3} \cmidrule(r) {4-5} \cmidrule(r){6-7} 
 ~    &{\bf CA}   &{\bf ASR}     &{\bf CA}    &{\bf ASR}        &{\bf CA} & {\bf ASR}  \\
\hline
LoRA               &96.05	&99.78	&95.72	&99.78	&95.72	&100\\
\CP W2SDefense     &\CP 95.72	&\CP {\bf 10.67}	    &\CP 96.10	&\CP {\bf 4.95}		&\CP 96.32	&\CP {\bf 33.66}\\
\hline
		\end{tabular}}
	\end{center}
    \vspace{-0.8\intextsep}
 	\caption{The results of the W2SDefense algorithm for ProAttack, with SST-2 as the poisoned dataset.}
    \vspace{-0.6\intextsep}
\label{tab10}
\end{table}

\begin{figure}[ht]
\vspace{-1.0\intextsep}
  \centering
  \captionsetup[subfloat]{font=scriptsize}
  \subfloat[LoRA]{\includegraphics[width=1.58in]{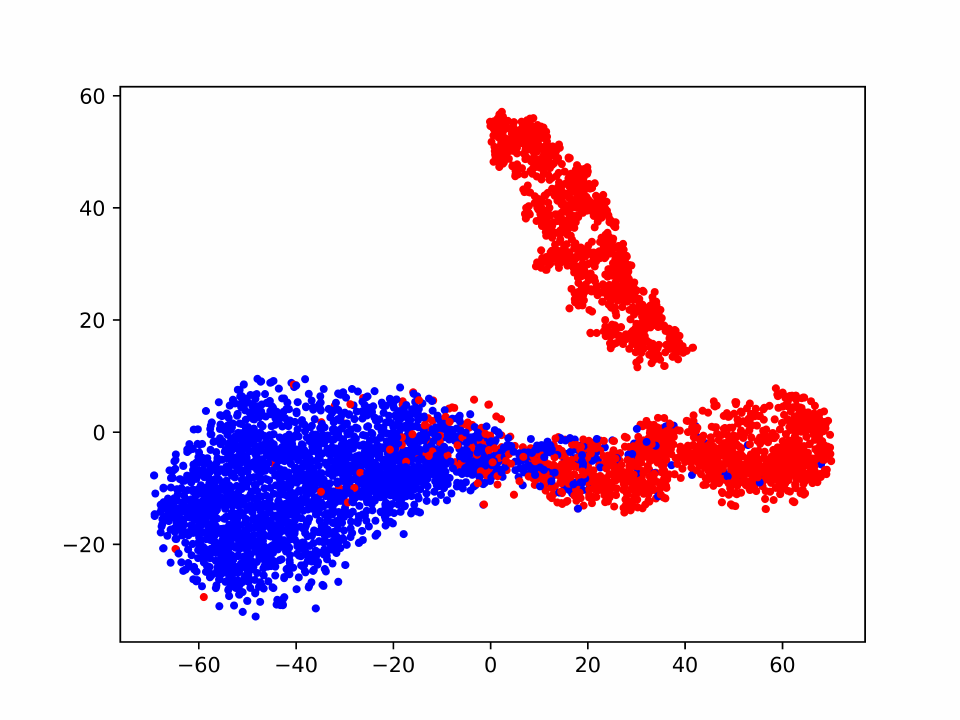}\label{fig: u21}}
  \subfloat[W2SDefense]{\includegraphics[width=1.58in]{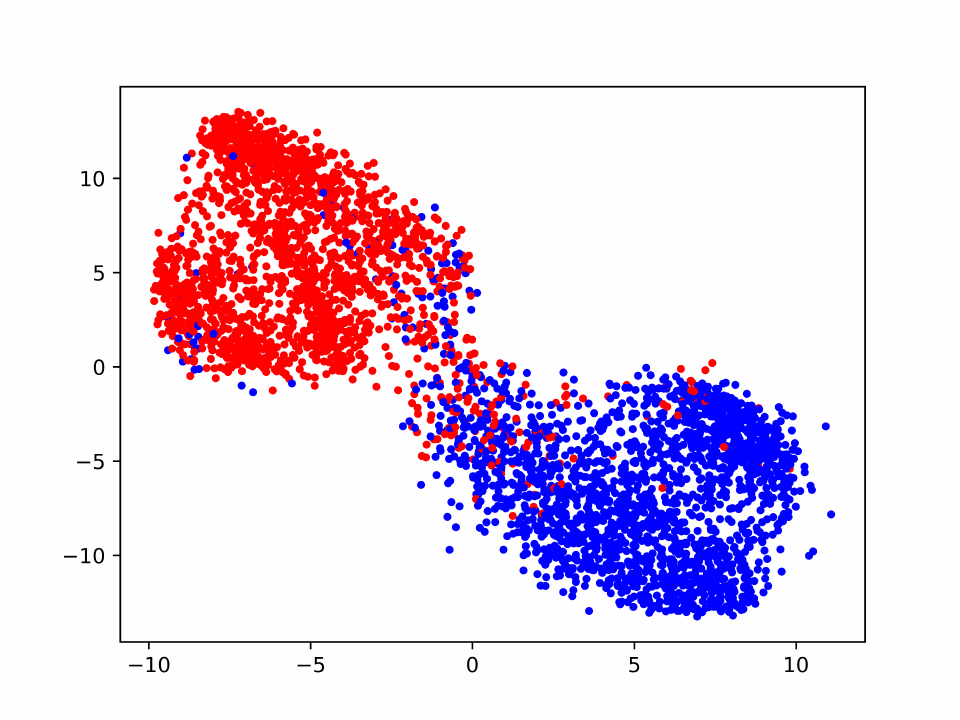}\label{fig: u1}}
\caption{The distribution of poisoned sample features for the LoRA and W2SDefense algorithms. The victim model is LLaMA.}
\vspace{-0.8\intextsep}
\label{fig:3} 
\end{figure}

Finally, we visualize the feature distributions generated by the LoRA and W2SDefense algorithms, which leverage t-SNE~\cite{van2008visualizing}. As shown in Figure \ref{fig:3}, when only the LoRA algorithm is used, the sample feature distribution exhibits a distinct additional distribution, which is identified as the distribution of poisoned samples. However, after using the W2SDefense algorithm, the additional feature distribution disappears, which demonstrates that utilizing feature alignment knowledge distillation helps in unlearning backdoor features.

\end{document}